\documentclass{egpubl}
\usepackage{EGVMV18}

\WsPaper         
\electronicVersion 

\ifpdf \usepackage[pdftex]{graphicx} \pdfcompresslevel=9
\else \usepackage[dvips]{graphicx} \fi

\PrintedOrElectronic

\usepackage{t1enc,dfadobe}

\usepackage{egweblnk}
\usepackage{cite}

\usepackage{booktabs}

\usepackage{subfigure}

\usepackage{amsmath}
\usepackage{siunitx}

\usepackage{xcolor}

\newcommand{\vect}[1]{\boldsymbol{\mathbf{#1}}}

\newcommand{\acknowledgments}[1]{
\section*{Acknowledgments}
#1}

\newif\ifhgt

\ifhgt
\newcommand{\change}[1]{
{\color{orange}#1}
}
\else
\newcommand{\change}[1]{
#1}
\fi

\title{Dynamic Environment Mapping for Augmented Reality Applications on Mobile Devices
\vspace{-15pt}}

\author[R. Monroy, M. Hudon \& A. Smolic]
{\parbox{\textwidth}{\centering R.\,Monroy\orcid{0000-0001-7472-0674}, M. Hudon and A. Smolic\orcid{0000-0001-7033-3335
}} \\
{\parbox{\textwidth}{\centering V-SENSE, School of Computer Science and Statistics \\ Trinity College Dublin, Dublin, Ireland}}
\vspace{-15pt}}

\begin{document}

\teaser{
  \includegraphics[width=\textwidth]{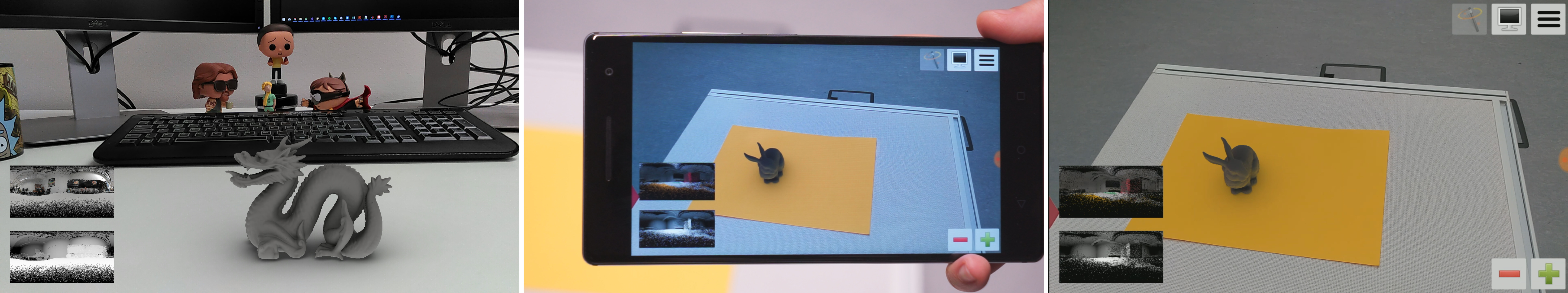}
  \caption{Sample results obtained with our pipeline. \textit{Left to right}: Resulting rendering of the Stanford Dragon after acquiring an office-space environment, capture of the mobile device from the user's point of view showing the Stanford Bunny, frame as seen directly on the device.}
}

\maketitle

\vspace{-5pt}
\begin{abstract}
    Augmented Reality is a topic of foremost interest nowadays. Its main goal is to seamlessly blend virtual content in real-world scenes. Due to the lack of computational power in mobile devices, rendering a virtual object with high-quality, coherent appearance and in real-time, remains an area of active research. In this work, we present a novel pipeline that allows for coupled environment acquisition and virtual object rendering on a mobile device equipped with a depth sensor. While keeping human interaction to a minimum, our system can scan a real scene and project it onto a two-dimensional environment map containing RGB+Depth data. Furthermore, we define a set of criteria that allows for an adaptive update of the environment map to account for dynamic changes in the scene. Then, under the assumption of diffuse surfaces and distant illumination, our method exploits an analytic expression for the irradiance in terms of spherical harmonic coefficients, which leads to a very efficient rendering algorithm. We show that all the processes in our pipeline can be executed while maintaining an average frame rate of 31Hz on a mobile device. 

\ccsdesc[500]{Human-centered computing~Mixed / augmented reality}
\ccsdesc[300]{Human-centered computing~Mobile devices}
\ccsdesc[500]{Computing methodologies~Mixed / augmented reality}

\printccsdesc  
\vspace{-5pt}
\end{abstract}

\section{Introduction}

\maketitle

The popularity of Augmented Reality (AR) applications is increasing everyday. The two biggest players in the mobile device arena have each released AR-oriented development kits for their respective platforms: ARKit (Apple) and ARCore (Google). In order to create convincing experiences, AR applications require the addition of virtual content that seamlessly blends with the real world, usually involving computationally-heavy rendering techniques. Some of these techniques are already able to be run in real-time in standard computers as shown by Purcell et al.~\cite{Purcell:2002:RTP}. \\
Every year faster and more efficient mobile devices are being released. However, even the fastest of these devices is far from matching the capabilities of a standard computer. These computing limitations motivate the need for new research efforts with the goal of creating realistic and interactive experiences capable of being deployed on today's mobile devices at interactive frame rates~\cite{Kronander:2015:PRM}. \\
In the present work, we propose a novel pipeline that allows for a coupled environment acquisition and virtual rendering in real-time on a mobile device. The system uses as input a collection of Low Dynamic Range (LDR) + Depth frames, commonly referred to as RGB-D images. When these images are combined with the pose of the device at the time of capture, they can be used to record the appearance of an environment, to which, virtual content is to be added. The description of the environment is stored as a two-dimensional image containing the environment's radiance plus depth for each pixel, which we refer to as the Environment Map (EM). Every time a new frame is acquired, the EM is updated and the virtual object is immediately rendered using the newest EM. By parallelising all the operations in our pipeline using the Graphics Processing Unit (GPU) of the mobile device, we achieve real-time frame rates ($>25Hz$). \\
Our main contributions are: $(1)$ an AR system capable of achieving real-time frame rates while simultaneously acquiring an environment and updating a virtual object's interactions with it, $(2)$ an image-based adaptive environment acquisition pipeline, (3) an efficient operator to relocate an EM and allow for a virtual object's translation.

\section{Related Work}

\begin{figure*}
    \centering
    \includegraphics[width=0.9\textwidth]{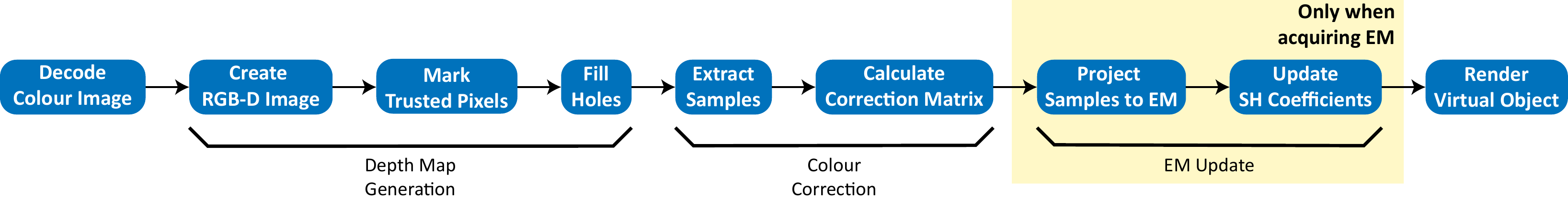}
    \vspace{-5pt}
    \caption{Overview of the different steps in our AR pipeline.}
    \label{fig:pipeline}
    \vspace{-15pt}
\end{figure*}

In general, most AR pipelines need to address four main topics: Simultaneous Localisation and Mapping (SLAM), colour correction, environment capture and rendering techniques. \change{In the presented work, we assume a relatively accurate pose is available and thus, no background on SLAM techniques will be discussed. For a good overview of some of these algorithms, we refer the reader to Cadena et al.~\cite{Cadena:2016:PPF}}.

\textbf{Colour correction} Most consumer products implement some variant of automatic exposure and white-balance. While this helps the casual creative user by removing the need to understand the effects of the parameters involved, it adds another level of uncertainty since the inner works are usually not publicly disclosed. Some methods to alleviate and even exploit this issue have been proposed. Kim and Pollefeys~\cite{Kim:2004:RAI} presented a technique to estimate the radiometric response function of a camera from a collection of images of the same scene under different exposures. This work was further extended by the same authors allowing for camera displacements while also correcting vignetting artefacts~\cite{Kim:2008:RRC}. Zhang et al.~\cite{Zhang:2016:ERR} proposed a global approach to estimate the camera response and create a High Dynamic Range (HDR) representation of an environment. In order to have instantaneous estimations, Rohmer et al.~\cite{Rohmer:2017:NEI} presented a method that progressively calculates a correction matrix that effectively transforms new frames into the colour space defined by a reference frame. In this work, we employ a technique similar to the one described by Rohmer et al., making use, however, of a different sampling strategy since the environment information in our system is stored as an EM.

\textbf{Environment Capture} In order to seamlessly render virtual content into a scene, knowledge about the environment is needed. Many methods have been proposed to capture lighting conditions in a scene. Debevec~\cite{Debevec:1998:RSO} proposed the use of a perfectly reflective sphere to capture an environment. This method was later used by Heymann et al.~\cite{Heymann:2005:IRR} to augment a video stream with a virtual object capable of reacting to changes in the scene lighting in real-time. Another technique is the use of one or several cameras with fish-eye lenses to reconstruct the lighting conditions as proposed by Havran et al.~\cite{Havran:2005:ISD}. \change{K\'{a}n et al.~\cite{Kan:2015:HQC} proposed a method that creates HDR EMs from a collection of pictures taken from a mobile device. Some other methods rely on 3D sensors to capture the effects of the illumination in a scene, this data is later stored as a mesh or an EM~\cite{Meilland:2013:3HDR, Rohmer:2017:NEI, Zhang:2016:ERR}. Wu et al.~\cite{Wu:2016:SLAE} proposed a method that uses an RGB-D system to estimate an object's appearance together with the environment illumination. More recently, Meta et al.~\cite{Meka:2017:LUGI} used a similar system, to determine the reflectance of surfaces, requiring some minor human interaction to solve inherent ambiguities. \\ 
In contrast with most of the aforementioned methods, we target the use of only resources available on mobile devices and no human input to acquire the EM. By expressing our EMs as RGB-D images, we find a middle ground between RGB EMs and 3D meshes, allowing some level of interactivity when moving the virtual object within the scene, all while keeping real-time performances.}

\textbf{Rendering} There are two main types of rendering techniques in AR as pointed out by Kronander et al.~\cite{Kronander:2015:PRM}: differential rendering and rendering based on EMs. \\
Differential rendering was first introduced by Fournier et al.~\cite{Fournier:1992:CIR} and involves the rendering of the scene twice: once only with the local model of the scene and another one adding the virtual object into the scene. Debevec~\cite{Debevec:1998:RSO} used this technique to augment scenes previously captured with a light probe. \\
Rendering based on EMs can be achieved by performing a Monte Carlo integration on each vertex to be rendered~\cite{Cook:1984:DRT}. However, if the scene illumination is of a high-frequency nature, a large amount of samples will be needed for an accurate representation~\cite{Pharr:2016:PBR}. Alternatively, lights could be detected in the EM and a virtual light could be placed in that position to later be used to create a corresponding shadow map. Meilland et al.~\cite{Meilland:2013:3HDR} use this last technique to project shadows on mirror-like virtual objects. Another method commonly employed is representing an environment through the calculation of Precomputed Radiance Transfer functions~\cite{Ramamoorthi:2001:ERI, Sloan:2002:PRT, Heymann:2005:IRR}. These functions essentially compress an EM using orthogonal basis functions. One type of such basis functions is the set defined by the Spherical Harmonics (SH) basis. Sloan et al.~\cite{Sloan:2002:PRT} demonstrated the advantages of using SH functions when used in the rendering process. \\
Due to the current hardware limitations on mobile devices, alternatives have been proposed to offload some of the processing tasks to a workstation, transmitting to the mobile device only critical data to render the virtual objects~\cite{Rohmer:2014:INF}. Other methods rely on a two-stage process: environment acquisition and rendering~\cite{Meilland:2013:3HDR,Zhang:2016:ERR,Rohmer:2017:NEI}. In this work we propose a method that captures an environment while rendering virtual content into the scene at the same time and in real-time, all this using solely the mobile device itself.

\section{Overview}

The proposed pipeline is currently implemented on a Tango-enabled device (Lenovo Phab 2 Pro). It can, however, be applied to any system with SLAM capabilities that captures RGB-D images. The Lenovo Phab 2 Pro delivers five point clouds per second containing a maximum of 38,528 points each, together with a FullHD LDR colour image. The point cloud and the colour image are provided with metadata containing attributes, such as their corresponding poses and distortions. By combining all this information, it is possible to generate an RGB-D frame \change{(224$\times$172 pixels)} which is the basic input for our system. \\
\change{Fig.~\ref{fig:pipeline} outlines the different processes involved in the pipeline. The colour camera delivers its data using the Y'UV420sp (NV21) format, which is decoded to obtain RGB data. The data obtained by the depth sensor is later merged with the RGB image to create a usable RGB-D image. This process is described in Section~\ref{sec:DepthMaps}. \\
Once an RGB-D frame is obtained, the corresponding pixels in the EM are found. When information from previous frames is available on the EM, these samples are paired to calculate a colour correction matrix that converts the colour space of the current frame to that of a reference frame, defined by the first frame captured. Section~\ref{sec:ColourCorrection} describes the steps needed to obtain this colour correction matrix. \\
The correction matrix is applied to the colour appearance of each RGB-D frame. After the correction, these pixels are projected to the EM, from which a new set of SH coefficients is calculated. Alternatively, it is possible to disable the EM update, e.g., if the EM has already been acquired. In this case the calculation of the correction matrix is the last step before rendering the virtual object. All these processes are discussed in Section~\ref{sec:EnvironmentCapture}. \\
The final step is the rendering of the virtual object into the scene using the newly-calculated SH coefficients. The specifics of how this is achieved are presented in Section~\ref{sec:Rendering}.}
\begin{figure*} 
    \centering
    \subfigure{\includegraphics[width = 0.2814\textwidth]{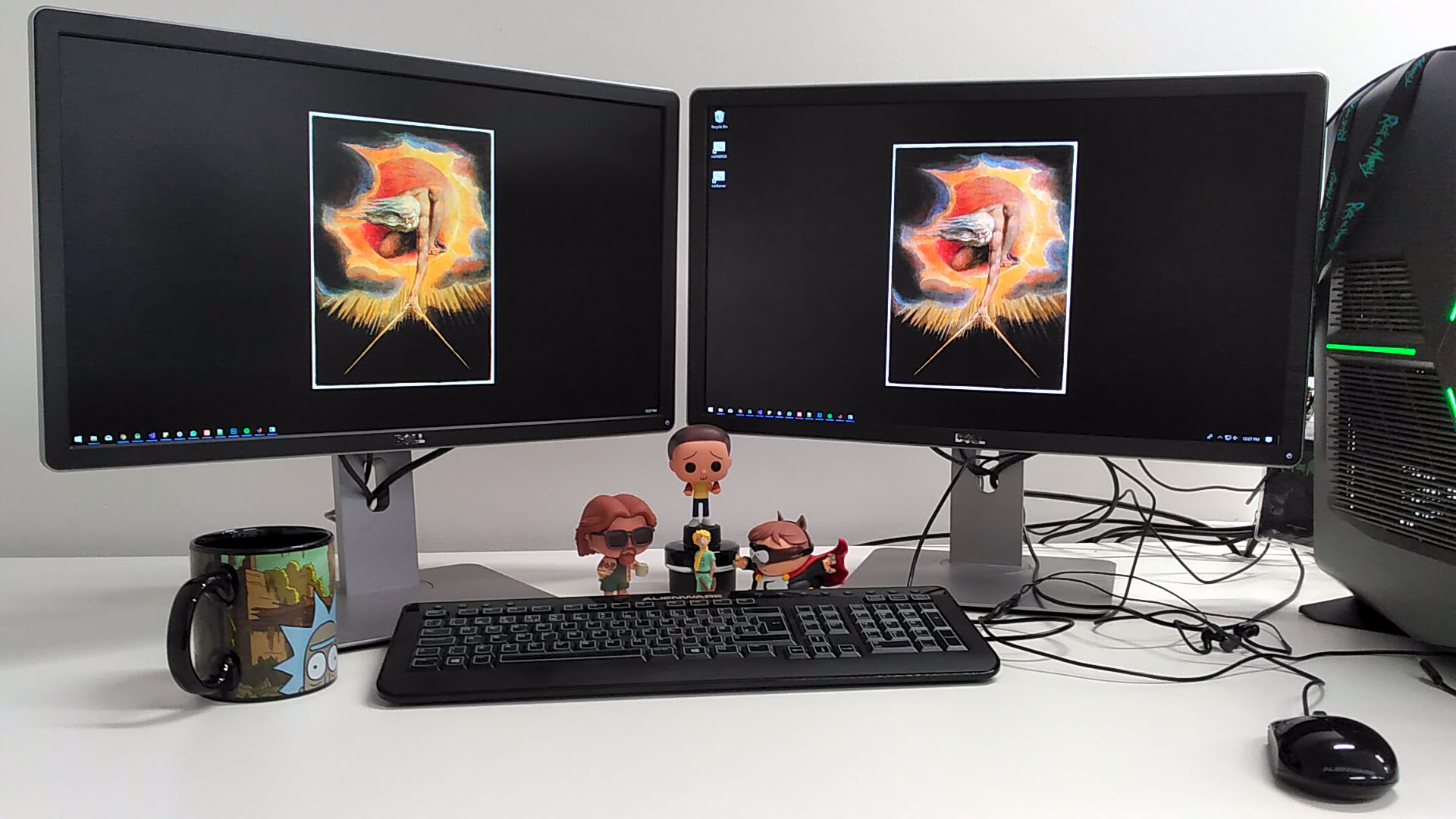}}
    \hspace{0.03\textwidth}
    \subfigure{\includegraphics[width = 0.2061\textwidth]{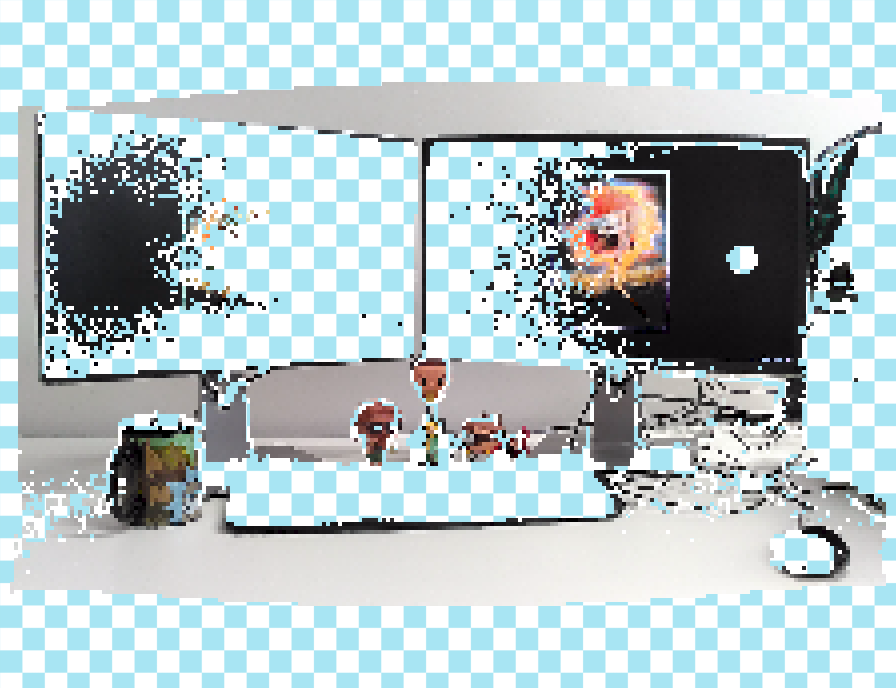}}
    \subfigure{\includegraphics[width = 0.2061\textwidth]{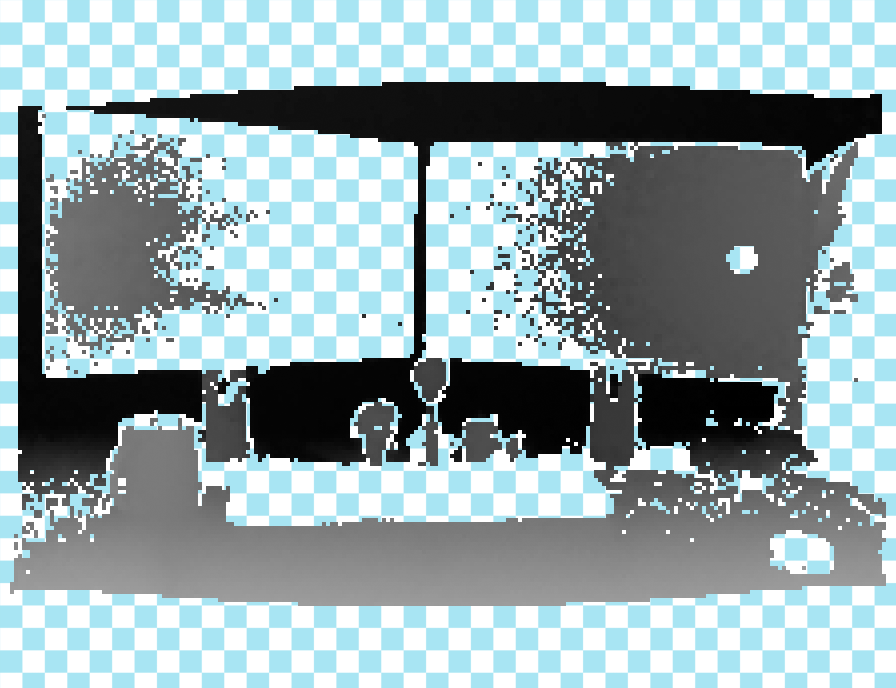}}
    \subfigure{\includegraphics[width = 0.2061\textwidth]{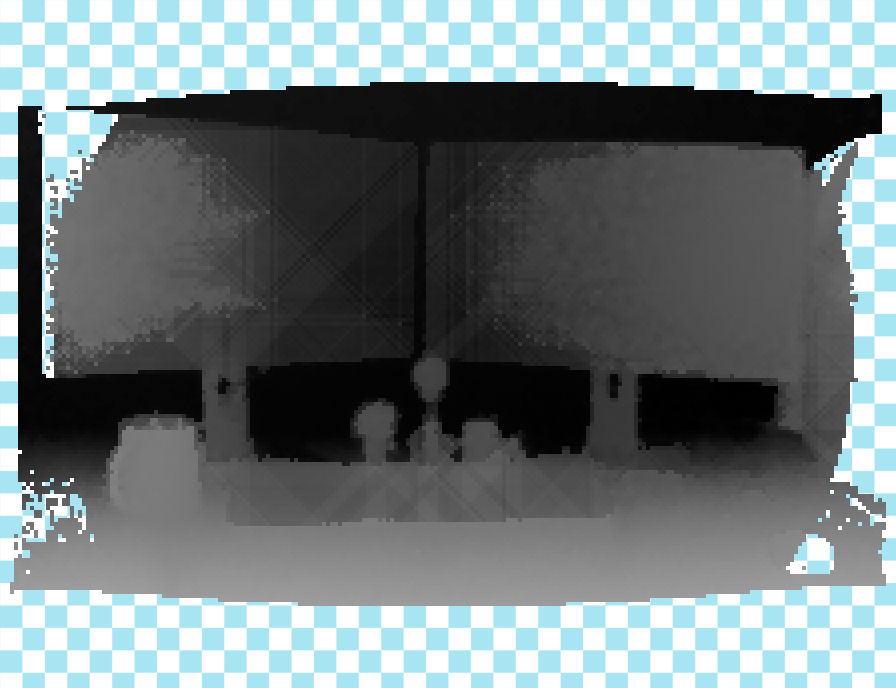}}
    \caption{Sample scenario showcasing missing data due to unreliable points in the depth map (Confidence value < 0.7), and how depth is estimated. \textit{Left to right}: Image from the colour camera, colourised depth map, depth maps before and after holes have been filled.}
    \label{fig:depthMapInit}
    \vspace{-15pt}
\end{figure*}

\section{Depth Maps}
\label{sec:DepthMaps}
The depth sensor on Tango devices operates using the Time-of-Flight (ToF) principle. \change{This sensor works reasonably well on most surfaces. However, there are some materials that are problematic and lead to unreliable measurements, e.g., shiny, translucent or light-emitting surfaces. Tango-enabled devices deliver depth data as a series of 3D points, each with a level of confidence between 0.0 and 1.0. Higher values indicate more reliable measurements.} \\
Tango defines so-called \textit{frames of reference} for the colour camera, the depth sensor and the real-world (global \textit{frame of reference}). These \textit{frames of reference} can be thought of as coordinate systems whose origins match those of the entities they belong to. This helps to interpret the position and orientation reported by the device for a given point cloud or colour image.

\subsection{Generating the RGB-D Image}
The colour camera and depth sensor exhibit some level of distortion, which is modelled using Brown's 3- and 5-Polynomial radial distortion models\cite{Tsai:1987:VCC}. The distortion coefficients and intrinsics matrices for both sensors are provided by Tango. With this data, it is possible to project points in the real-world to and from the different sensor planes: colour camera and depth sensor. \\
The first step towards creating an RGB-D image is finding the location of the corresponding pixel in the colour image. Since the point cloud uses the depth sensor's \textit{frame of reference}, it is necessary to apply the transformation between the colour camera and the depth sensor \textit{frames of reference}. Once the point is in the correct \textit{frame of reference}, we can apply the distortion coefficients and the intrinsic parameters of the colour camera to find the corresponding pixel in the colour image. Up to this point we are able to pair the points in the point cloud with their corresponding colour. To find the pixel location for each point on the RGB-D frame, we then use the distortion coefficients and intrinsic parameters of the depth sensor. Since the RGB-D image shares the same \textit{frame of reference} as the depth sensor, there is no need to apply a transformation to the points. \\
Fig.~\ref{fig:depthMapInit} on the extreme right, illustrates the resulting RGB-D image when acquiring a scene with computer screens present. In this particular case, only points with a confidence value $\geq$ 0.7 are considered reliable, the rest are not visualised (checkered pattern).

\subsection{Filling Holes in the RGB-D Image}
As can be seen from Fig.~\ref{fig:depthMapInit}, the surfaces of the computer screens and the keyboard are not accurately acquired. The same situation occurs with windows and most light-emitting surfaces. Rohmer et al.~\cite{Rohmer:2017:NEI} proposed a method that allows the user to manually trigger the addition of such surfaces into their environment representation. In order to reduce the need for human interaction and improve the efficiency of our pipeline, we estimate the missing depth values from reliable pixels in the neighbourhood. For this, we look for the closest reliable pixels along the eight cardinal and intercardinal directions on the RGB-D frame. The final estimated depth value $d$ is the weighted mean of the reliable values found:
\begin{equation}
\label{eq:depthEstimation}
d = \frac{\sum_{n=1}^{8} w_n d_n}{\sum_{n=1}^{8} w_n}
\end{equation}
where 
\begin{math}
w_n = \frac{w - x_r}{w}
\end{math}. The distance to the closest reliable pixel holding depth $d_n$ is represented by $x_r$. The width of the RGB-D image is denoted by $w$. When a direction leads to no reliable pixel, $w_n$ takes the value of zero. If none of the directions returned a reliable pixel, the pixel will remain unknown. The last two images in Fig.~\ref{fig:depthMapInit} show a comparison of the depth maps before and after this operation. Even though some of the estimated depth information might not be very accurate, it is usually sufficient to create a plausible EM from it. \\
The colours stored in the RGB-D image are linearised using the transformation defined by the sRGB colour space standard~\cite{IEC61966-2-1:1999}, which was verified using an X-Rite ColorChecker. This transformation provides a slightly better linear fit compared to using a $\gamma = 2.2$ as suggested by Zhang et al.~\cite{Zhang:2016:ERR}.

\section{Colour Correction}
\label{sec:ColourCorrection}
In order to properly insert virtual content that blends realistically into a scene, a colour correction needs to be performed on the appearance of the virtual object such that it matches the most up-to-date exposure and white-balance on the live-feed image.

\subsection{Colour Correction Matrix}
The problem of finding a colour correction transformation can be formulated as a non-linear optimisation problem as proposed by Zhang et al.~\cite{Zhang:2016:ERR}, where they look for three per-channel and independent factors used to model the effects derived from changes in exposure and white-balance. However, this optimisation is performed globally once an entire scene has been acquired. Rohmer et al.~\cite{Rohmer:2017:NEI} proposed instead, an approach that calculates a colour correction matrix to transform each new frame to a reference colour space. This reference can be defined, for instance, as the colour space used in the first captured frame. We apply a similar approach adapting only the sampling strategy to fit the characteristics of our RGB-D EM, as is described below. \change{It is worth mentioning that both of these approaches suffer from a scale ambiguity, which is, to some degree, alleviated by using the first frame as a reference.}

\subsection{Collecting the Paired Samples}
\begin{figure}
    \centering
    \subfigure{\includegraphics[width = 0.23\textwidth]{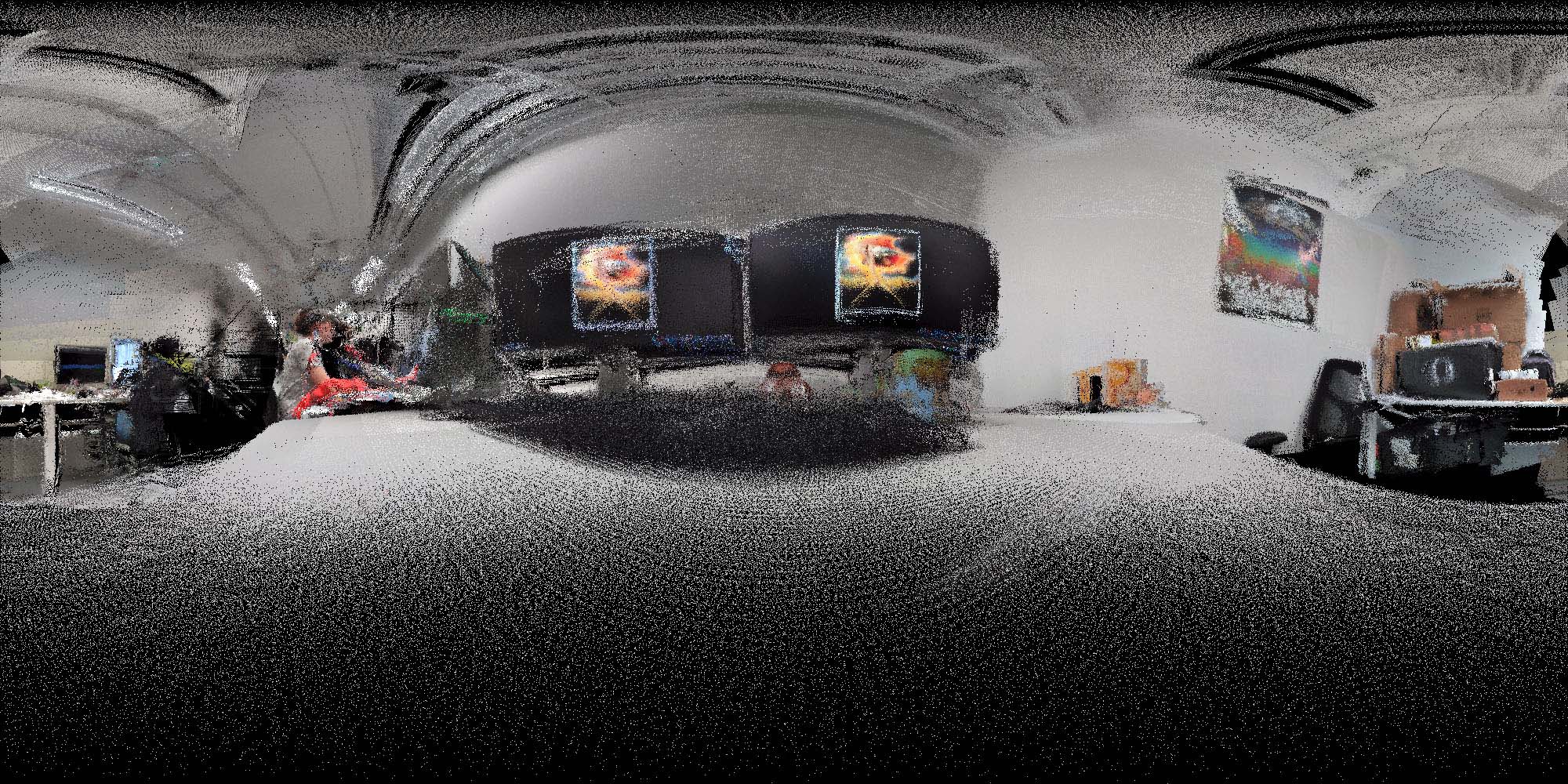}}
    \subfigure{\includegraphics[width = 0.23\textwidth]{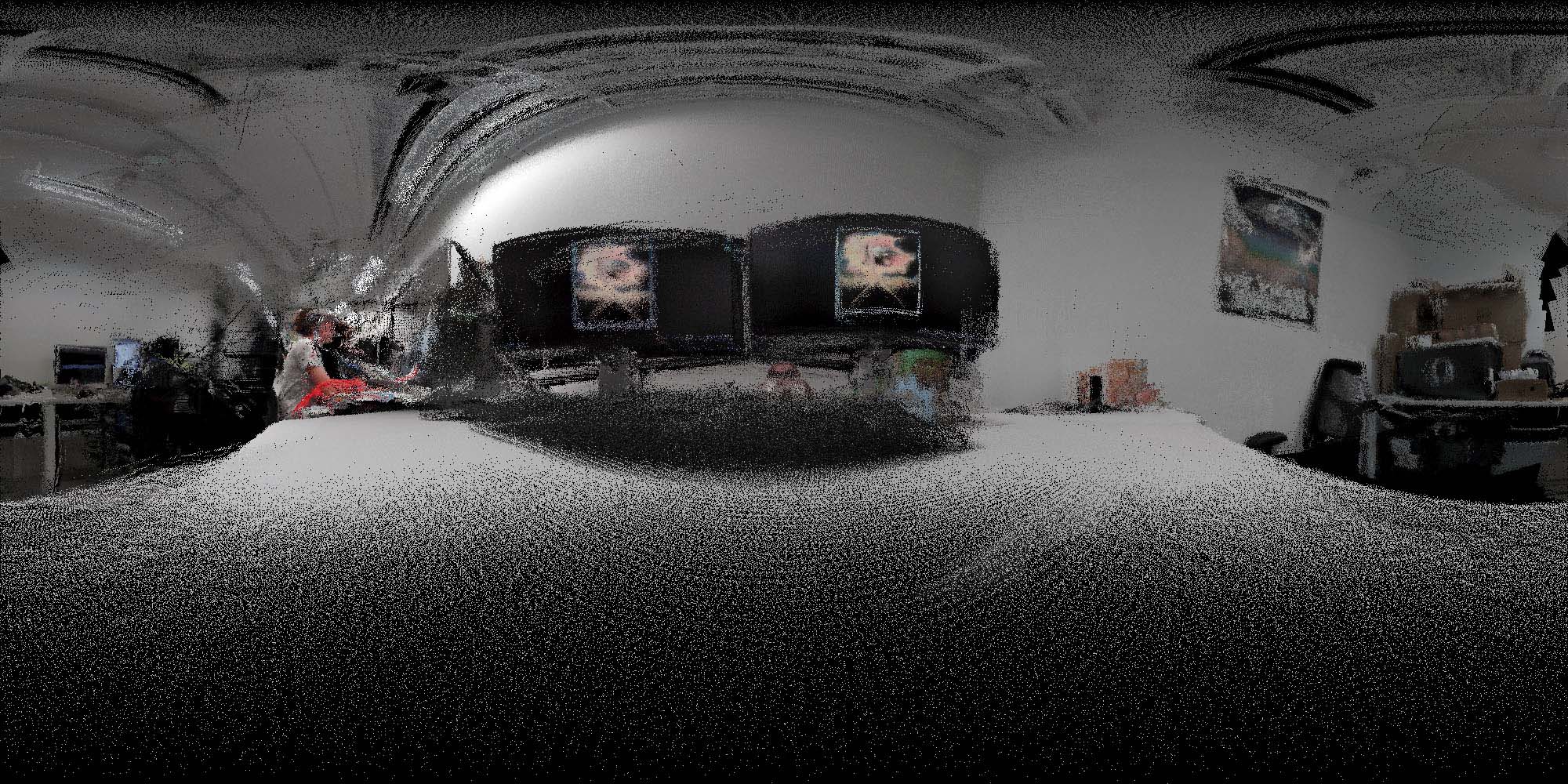}}
    \caption{Comparison of the same scene with colour correction disabled \textit{(left)} and enabled \textit{(right)}.}
    \label{fig:emComparison}
    \vspace{-15pt}
\end{figure}

In order for the colour correction process to deliver acceptable results, we need to make sure the paired sample points describe the same surface in the scene. We label all pixels in the RGB-D image as reliable or unreliable and use only reliable points when calculating the correction matrix. This labelling operation is performed before filling the holes in the RGB-D image. We first reject under- and over-exposed pixels, which we define as pixels with all their RGB values below 0.05 or above 0.95 respectively (in a normalised range from 0.0 to 1.0). Points that are close to depth discontinuities are also discarded. We do this by performing a $\chi^2$-test on the 8-connected pixels around each known point as proposed by Mitsunaga and Nayar~\cite{Mitsunaga:1999:RSC} to find flat surfaces. In our case, we use only depth information to perform this test. \\
We find the matching pixel in the EM by transforming a point from the depth sensor's \textit{frame of reference} to the EM's coordinate system, defined by its origin in the global \textit{frame of reference}. Finally, we use an equirectangular projection to find the corresponding pixel on the EM. Once the paired sample data is collected, points for which we do not have reference data are discarded. Additionally, points with large depth differences are also discarded. Since the accuracy of the depth sensor ranges between 0.1cm and 4cm depending on the measurement, we define this threshold as 1cm as a middle ground. This trusted set of paired samples is then used to calculate the colour correction matrix. \\
After computing the colour correction matrix we calculate its corresponding Mean Squared Error (MSE) from the pairs of trusted samples. If the MSE is above a certain threshold, we revert to using the previously accepted correction matrix for correcting the colour image but the samples are not projected to the EM. Following experimentation, we set the threshold for the MSE as \num{5e-2}. Fig.~\ref{fig:emComparison} provides a comparison between two EMs created from the same input with the colour correction disabled and enabled. Notice the difference in luminance and colour drift particularly on the wall directly under the light and around the computer screens.

\section{Environment Capture}

\label{sec:EnvironmentCapture}
The EM is an RGB-D image that maps the radiance in an environment coming from all directions onto a defined centre, which we refer to as the EM's origin, and is related to the position of the virtual object. In our current implementation this EM has a size of 2000$\times$1000 pixels. Smaller sizes can cause problems when computing the colour correction matrix because surface points with potentially different colours might be mapped onto the same pixel. The chosen size is a compromise between accuracy and speed. \\
The creation of the EM is a relatively straight-forward step, we only need to define a set of rules on how and when its content is updated. At this point the colours of the current RGB-D image are adjusted by applying the correction matrix. The corresponding pixel location in the EM of a given point on the RGB-D image is obtained by projecting it onto the EM. 

\subsection{EM Update}
\begin{figure}
    \centering
    \includegraphics[width=0.45\textwidth]{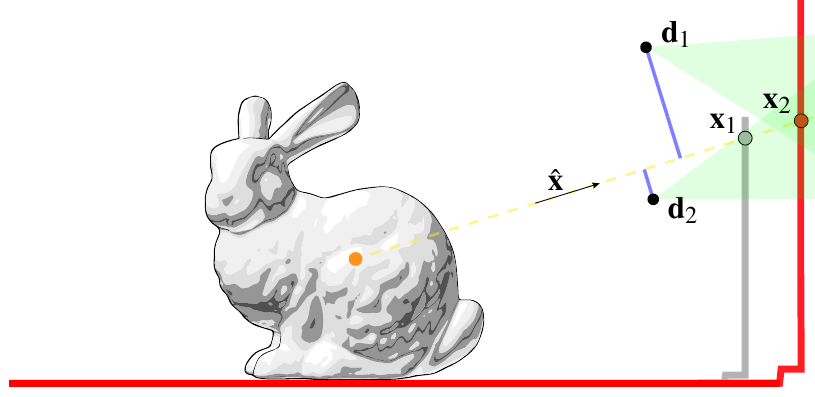}
    \caption{EM acquisition from two different device locations $\textbf{d}_1$ and $\textbf{d}_2$. Only data from $\textbf{d}_2$ can be used to remove point $\textbf{x}_1$.}
    \label{fig:emCase3}
    \vspace{-15pt}
\end{figure}

\change{The EM is updated keeping data from the closest objects to the EM's origin. In order to allow for changes in the scene, there is one case that needs special attention: when an object is removed. This kind of update is only triggered when the device's location is between the previously-observed surface and the EM's origin. Fig.~\ref{fig:emCase3} illustrates this situation. The grey surface containing point $\textbf{x}_1$ belongs to an old EM, i.e., an object that was removed. The red surface containing point $\textbf{x}_2$ represents a new observation. Notice that both points, $\textbf{x}_1$ and $\textbf{x}_2$, would be mapped to the same pixel since the normalised vector $\hat{\textbf{x}}$, from the EM's origin (orange point) to each point, is the same. \\
Let us analyse the case in which $\textbf{x}_2$ is observed from two locations $\textbf{d}_1$ and $\textbf{d}_2$ (sensor's frustums are shown as green triangles). Only the observation from $\textbf{d}_2$ can be used to remove $\textbf{x}_1$ because this point cannot be seen from $\textbf{d}_1$, and thus, it is not possible to confirm it must be removed based on the data obtained from this vantage point. We need to make sure the sensor's frustum contains the surfaces that are to be removed. This is implemented by measuring the distance from the device's position to the closest point on $\hat{\textbf{x}}$ (blue line). If the distance is smaller than a predefined threshold, we allow an update on that pixel. Following experimentation, this threshold is set as 3cm. Additionally, pixels belonging to a surface's backface are detected and ignored. This check is performed by measuring the projected distance of the vector from the EM's origin to the device position on $\hat{\textbf{x}}$. If the distance is larger than the depth from the previous or current observation for that pixel, it belongs to a backface. \\
We also allow a swift override of all the checks by defining a \textit{trusted volume}, delimited by a sphere centred on the EM's origin. If the device is inside, i.e., close to the optimal location to create the EM, the previous data is immediately replaced. In our implementation this sphere has a radius of 10cm. \\
When updating a pixel in the EM, the previous colour data is immediately replaced. The depth data, on the other hand, is updated using the mean value of the previous and current depths. The reason for using the mean value, is reducing the impact of inaccurate depth estimations, e.g., derived from the hole-filling operation. All these rules have the combined effect of allowing small changes in the scene as demonstrated in the supplementary video.}

\subsection{EM Translation}
\label{sec:transEM}
\begin{figure}
    \centering
    \subfigure{\includegraphics[width = 0.23\textwidth]{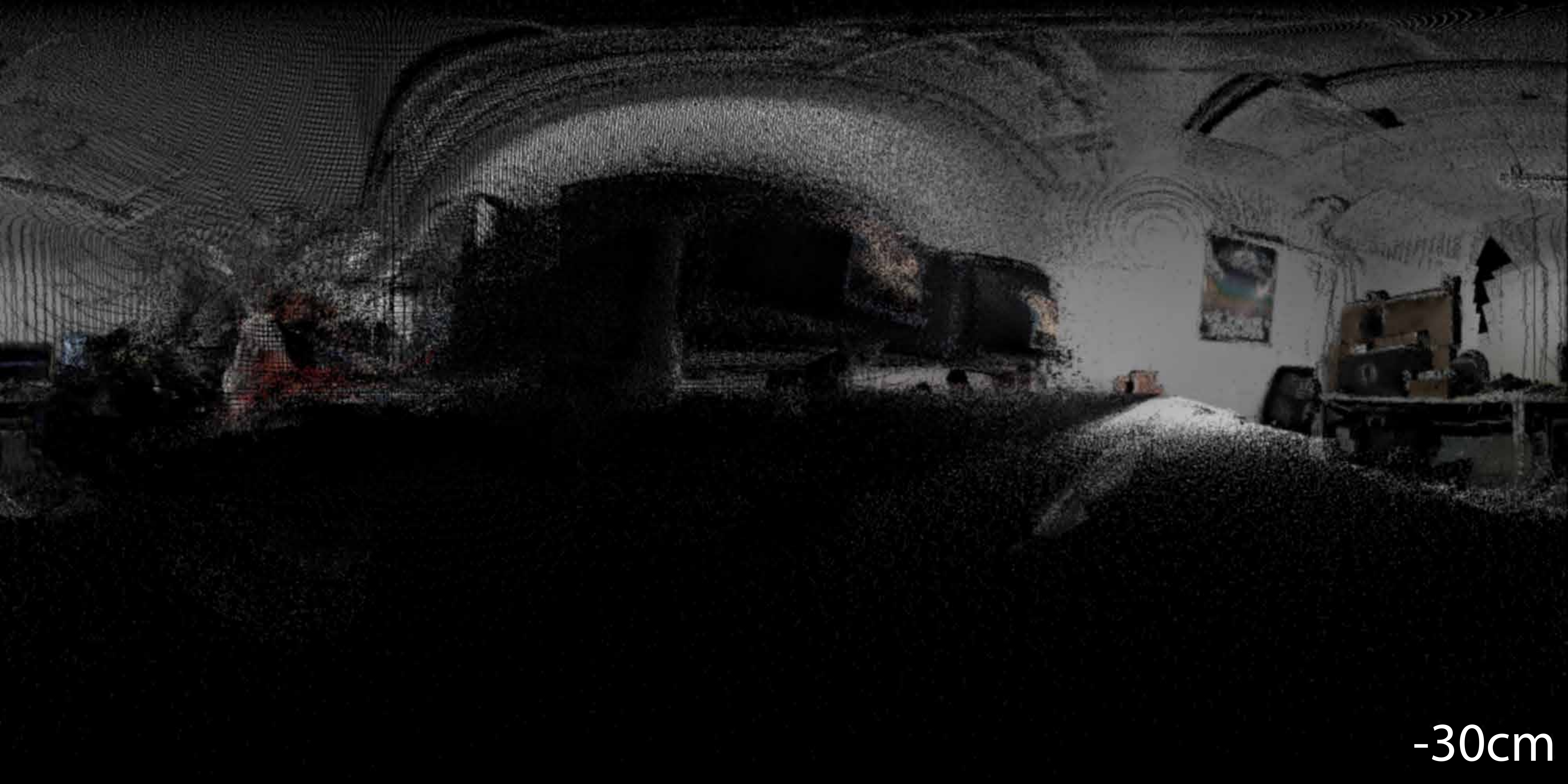}}
    \subfigure{\includegraphics[width = 0.23\textwidth]{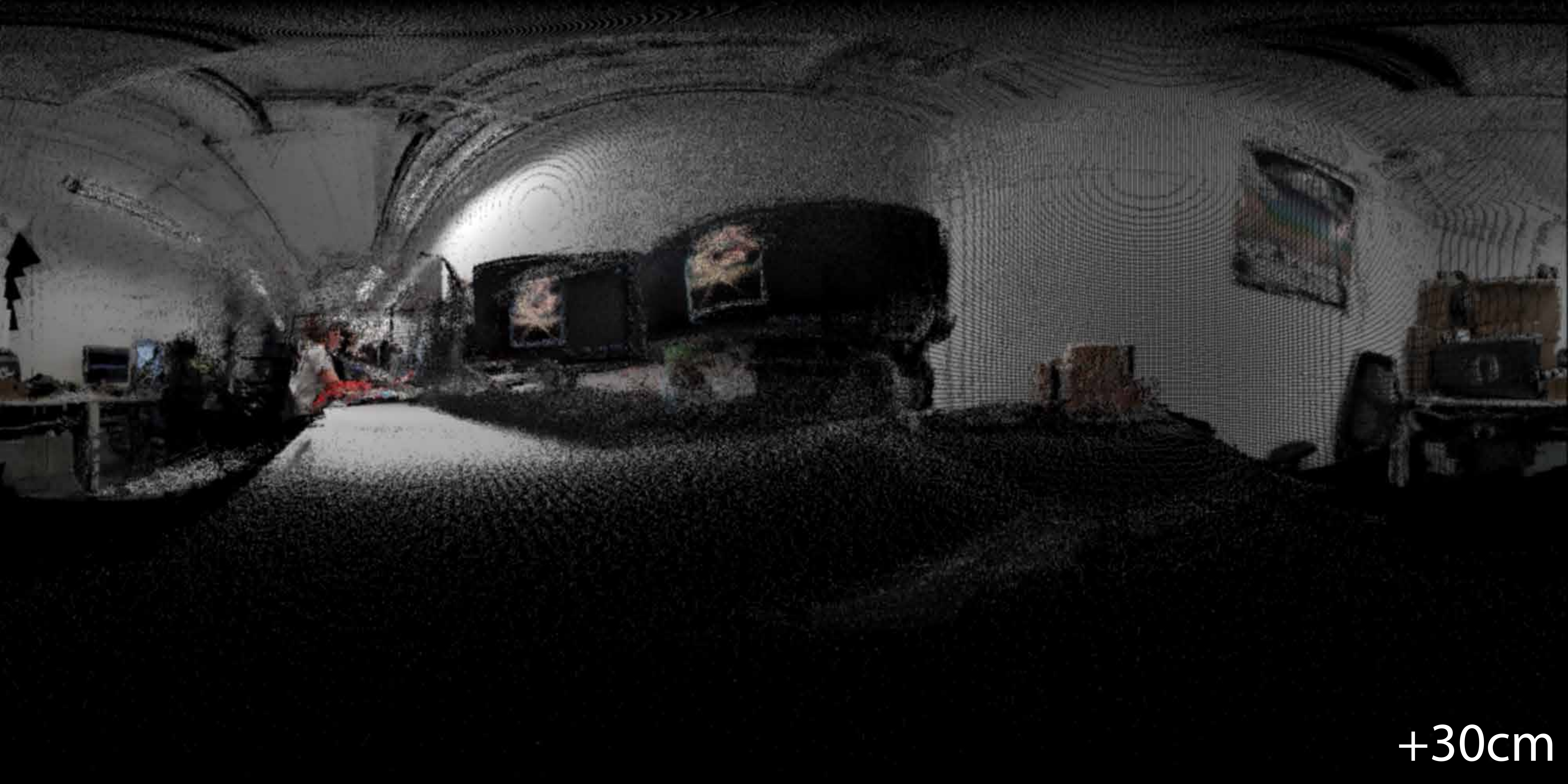}}
    \caption{EM displacements along the x-axis.}
    \label{fig:emDisplacement}
    \vspace{-15pt}
\end{figure}

\change{Gardner et al.~\cite{Gardner:2017:LPII} proposed a warping operator that simulates EM translations. This operator works under the assumption that all surfaces on the EM are far away, which is usually not the case, especially if local effects of objects nearby are expected to be simulated. When objects are present close to the EM's origin, displacements are excessively exaggerated.} \\
Since depth information is stored in the EM, it is possible to recover the three-dimensional position of the pixels in the EM within the global \textit{frame of reference}. Once the global position of the points is recovered, the points are projected using the new and translated EM's origin\change{and their corresponding depths recalculated.} Fig.~\ref{fig:emDisplacement} illustrates the effect of displacing the EM shown in Fig.~\ref{fig:emComparison} by 30cm in both directions along the x-axis. \\
The new EM can overwrite the values on the initial EM or be kept as a temporary EM instead. The first option allows the addition of RGB-D data from the new EM's origin, while the second one enables continuous motions without losing information. This last option will map new RGB-D data using the initial EM's origin.

\section{Rendering}
\label{sec:Rendering}

As shown by Kajiya~\cite{Kajiya:1986:RE}, the incoming radiance received on a surface can be calculated by integrating the received light from all the directions in a hemisphere above the surface. For perfectly Lambertian surfaces the integral is reduced to:
\begin{equation}
\label{eq:renderingEquation}
L(\textbf{x}) = \frac{\rho}{\pi} \int_{\Omega}L_{i}(\textbf{x}, \vect{\omega}{}_i)(\textbf{n} \cdot \omega{}_i)d\omega{}_i
\end{equation}
where $L$ is the outgoing radiance at point $\textbf{x}$. The albedo of the surface is represented by $\rho$, while $L_i$ denotes the environment's radiance from the direction $\vect{\omega}{}_i$. The normal of the surface at point $\textbf{x}$ is represented by $\textbf{n}$.

\subsection{Spherical Harmonics}
Solving Equation~\ref{eq:renderingEquation} can be considerably expensive. The most common approach involves limiting the integration to a set of random directions, i.e., Monte Carlo integration. Sloan et al.~\cite{Sloan:2002:PRT} demonstrated how SH offer a very efficient alternative. This technique compresses a spherical function by projecting it using the SH basis functions, resulting in a series of coefficients. Increasing the number of SH bands used (more coefficients), leads to a more accurate approximation. Furthermore, if two spherical functions are projected onto their SH coefficients, and the integral of their product is desired, this can be approximated by calculating the dot product of their coefficients due to the orthonormality of the SH basis functions. We exploit this property by setting these spherical functions as: the radiance distribution of the environment; a spherical function encoding the per-vertex visibility due to the self-occlusion of the virtual object and the cosine-weighted effect of incoming light from a given direction. The coefficients of the latter function are precomputed once per virtual object. \\
It is not uncommon to observe seemingly large portions missing in the EM, as can be seen in Fig.~\ref{fig:emComparison} on the lower area. This does not necessarily indicate a lack of information, but is a side-effect of the distortions caused by the equirectangular projection and the presence of nearby surfaces. To alleviate this problem, when computing the SH coefficients of the EM, if the sampled pixel on the EM has no data, the colour of the closest 4-connected pixel is used. We limit the search to the neighbouring 50 pixels in each direction, this was found to be sufficient to address this issue. If the EM acquisition is executed continuously, over time most of these missing pixels will be filled in. If no known data is found, the sample is skipped when calculating the coefficients. \\
In order to simulate soft shadows in the scene, we create a plane under the virtual object, for which we precompute the SH coefficients at several locations encoding the visibility changes due to the virtual object and Lambert's cosine law. This object is rendered as a dark-grey plane whose alpha value is indicated by the difference of estimated irradiance with and without the virtual object.

\subsection{Compositing}
The depth sensor delivers a new point cloud every 200ms. However, the colour camera provides a new image with a much higher frequency allowing for real-time interactions. In order to keep interactive frame rates, we correct the colour appearance of the virtual object with the inverse of the latest accepted correction matrix, which is calculated every time a new point cloud is available. Applying this matrix ensures a consistent experience, even when the automatic white-balance and exposure change. \\
On every rendering cycle, the colour image is shown on the background, and the virtual content is rendered using a virtual camera that matches the perspective of the colour camera. In the current implementation no depth-check is performed since it is not the main focus of the presented work, but could be added to add further realistic effects due to the occlusion of objects in the real-world.
\begin{table}
\centering
\caption{Timings for the processes in the proposed pipeline. Steps marked with * are performed only while the EM is recorded. }
\label{tab:timeProc1}
\vspace{-5pt}
{\tiny
\begin{tabular}{lr}
\textbf{Process}                 & \multicolumn{1}{l}{\textbf{Elapsed time [ms]}}  \\ 
\toprule
Transfer data to GPU             & 10.02                                            \\
YUV420 to linear RGB             & 1.00                                            \\
Point cloud to RGB-D             & 2.69                                            \\
Mark reliable RGB-D              & 0.06                                            \\
Hole filling RGB-D               & 0.08                                            \\
Collect paired samples           & 2.35                                            \\
Compute colour correction matrix & 13.97                                           \\
Calculate MSE                    & 2.69                                            \\
Project sample to EM*            & 0.21                                            \\
Calculate SH coefficients*       & 0.46                                            \\
Render frame                     & 20.7                                           \\
\hline
\textbf{Total}                   & \textbf{54.22}                                             
\end{tabular}
}
\vspace{-10pt}
\end{table}

\section{Results and Discussion}

In this section we present results obtained in real-world scenarios when applying our proposed pipeline. All of them are based on an implementation of the pipeline running on the aforementioned Lenovo Phab 2 Pro, which features a Qualcomm Snapdragon 652 system-on-chip with an Adreno 510 GPU. \change{The pixel values in each of the steps of the pipeline depend only on previously computed data, and thus, each stage was implemented as a compute shader. Most of these shaders have no pixel interdependency and were implemented in a straight-forward manner. The calculation of the correction matrix and the SH coefficients, however, involve a large amount of sums, which are arranged as prefix sums to efficiently exploit the GPU capabilities. The details of the shader implementation for these two steps are presented in the supplemental material. The SH rendering was implemented as standard pair of fragment and vertex shaders.} For further demonstrations, we refer the reader to our accompanying video. 

\textbf{Performance Timing} In our implementation the time required for all the processes illustrated in Fig.~\ref{fig:pipeline} is 54.22ms on average. However, since the pipeline is only run in its entirety whenever new depth data is available (every 200ms), the actual average elapsed time considering all the frames that are rendered is 31.74ms, leading to an average frame rate of 31.51Hz. These values were obtained by averaging the timings of over more than 1,000 rendered frames while recording and visualising the Stanford bunny ($>$34, 000 vertices). The individual timings per process are shown in Table~\ref{tab:timeProc1} and correspond to renderings using five SH bands. \\
It is worth mentioning that when the recording of the EM is not required, e.g., the scene is static and has already been captured, the EM and its SH coefficients are not updated and thus, those processes are not executed. However, the impact on the final elapsed time is minimal.\\
The number of SH bands can be increased, allowing higher-frequency content in the scene to be integrated. However, this dramatically affects the average frame rate, mainly because of the additional computations per vertex when rendering the virtual object. We observed that having more than five bands did not led to noticeable differences in the final render. As mentioned by Sloan et al.\cite{Sloan:2002:PRT}, having four or five bands is enough for typical meshes and low-frequency environments.\\
The most time-consuming process in the pipeline is the estimation of the colour correction matrix, as can be seen in Table~\ref{tab:timeProc1}. However, if manual control over the exposure and white-balance were available, this step would not be necessary and frame rates of about 40Hz could be achieved.

\begin{figure}
    \centering
    \subfigure{\includegraphics[width = 0.23\textwidth]{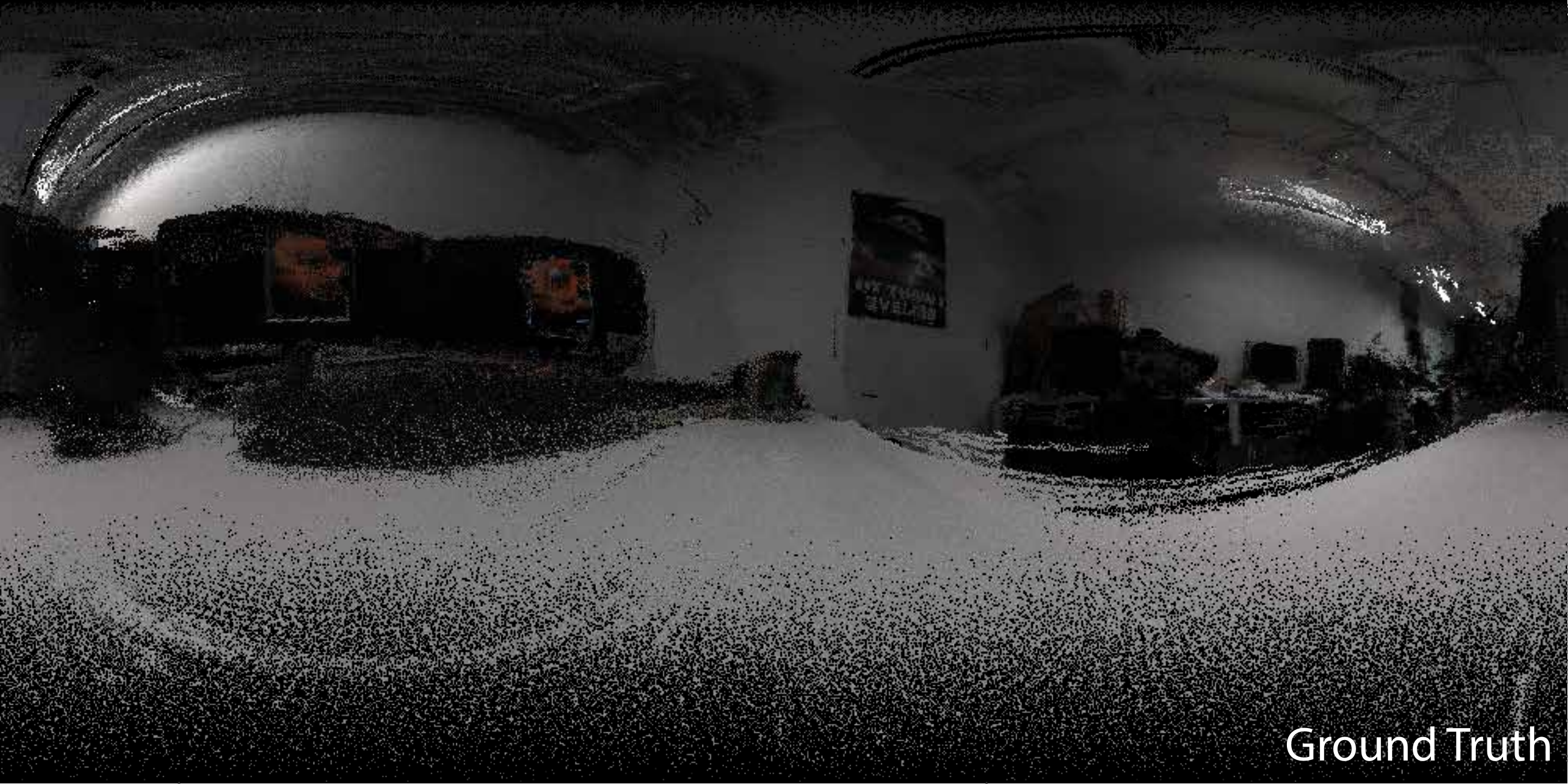}} \\
    \vspace{-10pt}
    \subfigure{\includegraphics[width = 0.23\textwidth]{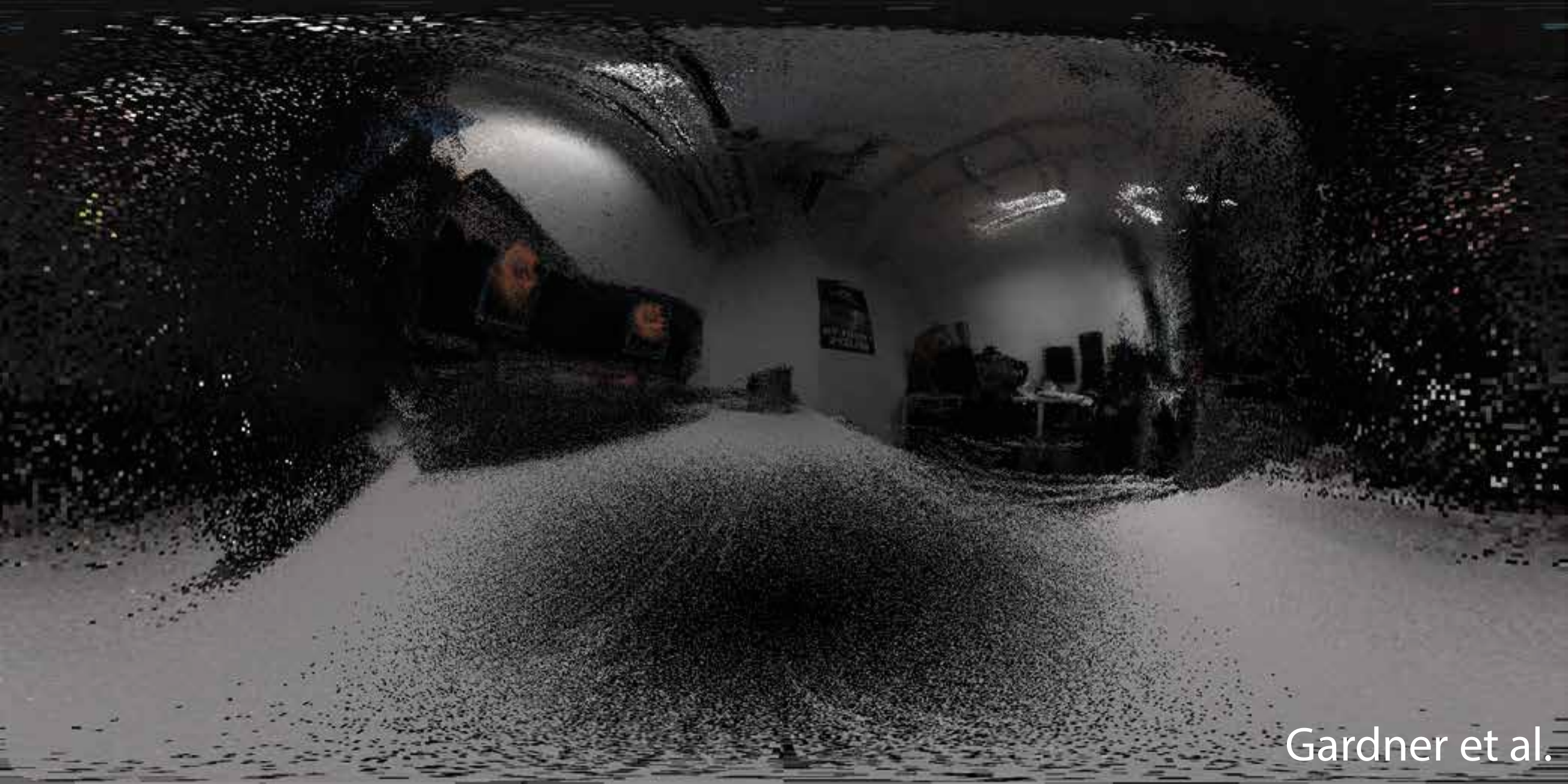}}
    \subfigure{\includegraphics[width = 0.23\textwidth]{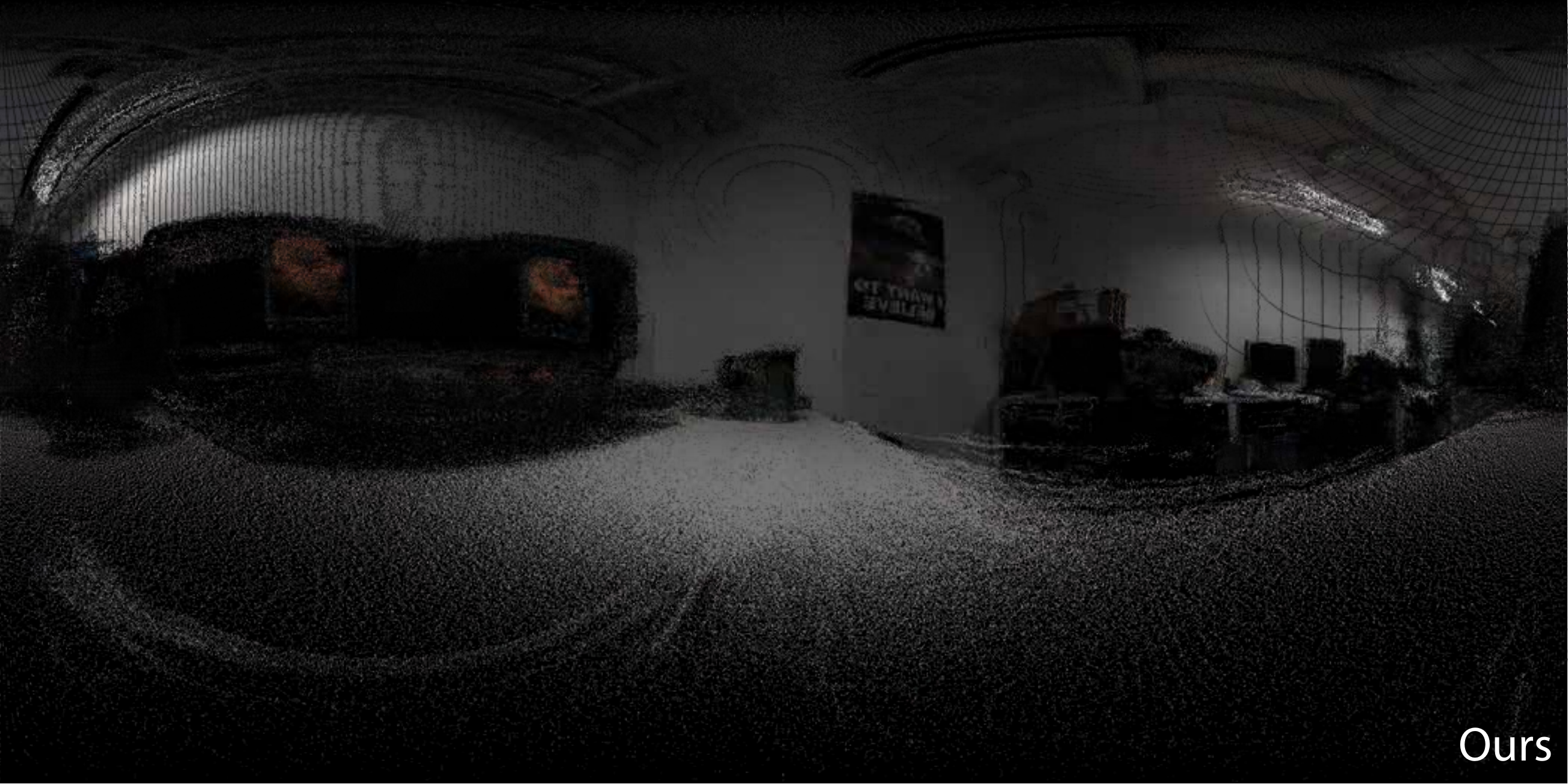}}
    \vspace{-5pt}
    \caption{EM Translation comparison. \textit{Above}: Ground truth. \textit{Below}: Using Gardner et al. operator (\textit{left}) and ours (\textit{right}).}
    \label{fig:transComparison}
    
    \vspace{-10pt}
\end{figure} 

\begin{figure}
    \centering
    \subfigure{\includegraphics[width = 0.23\textwidth]{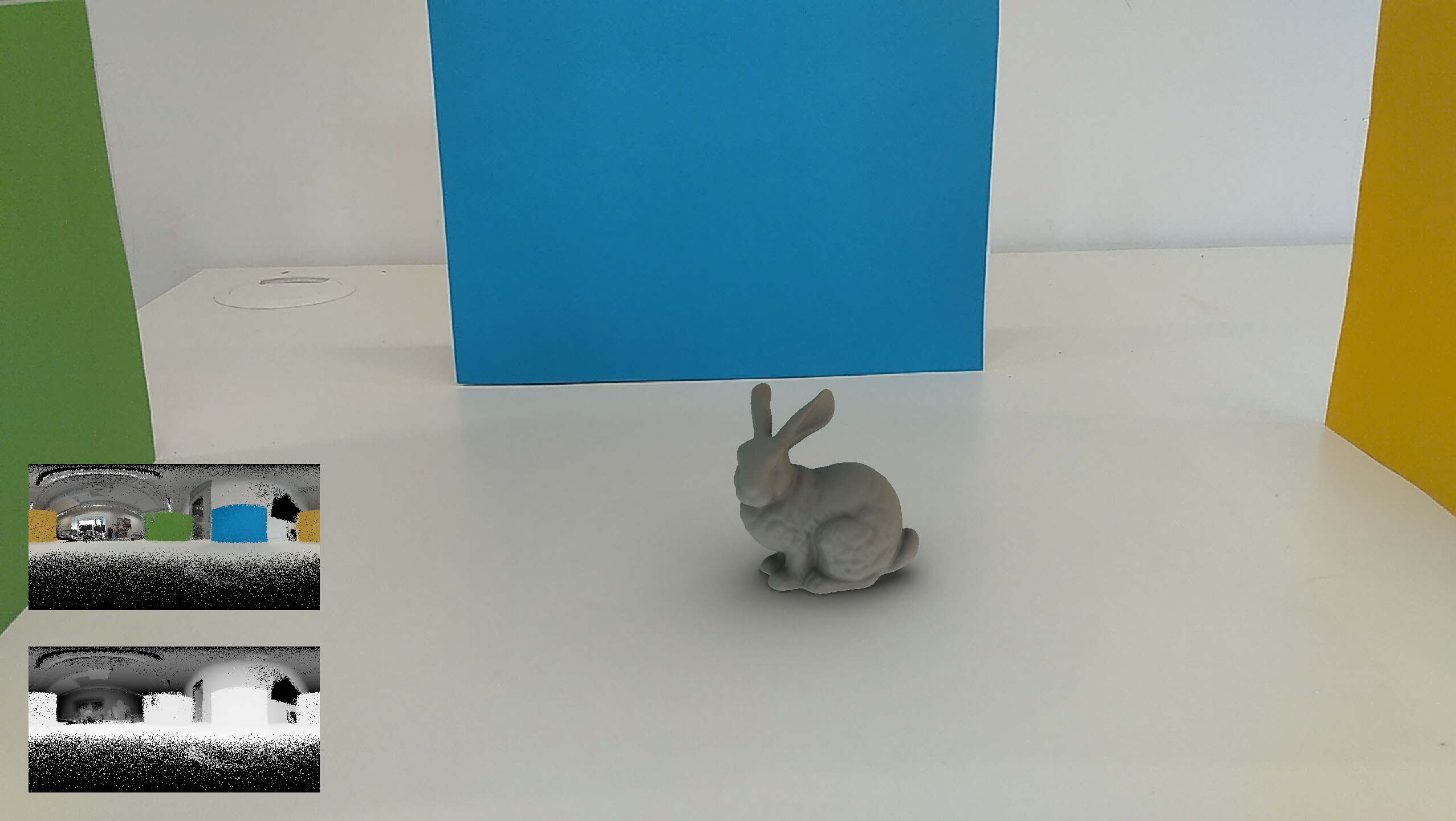}}
    \subfigure{\includegraphics[width = 0.23\textwidth]{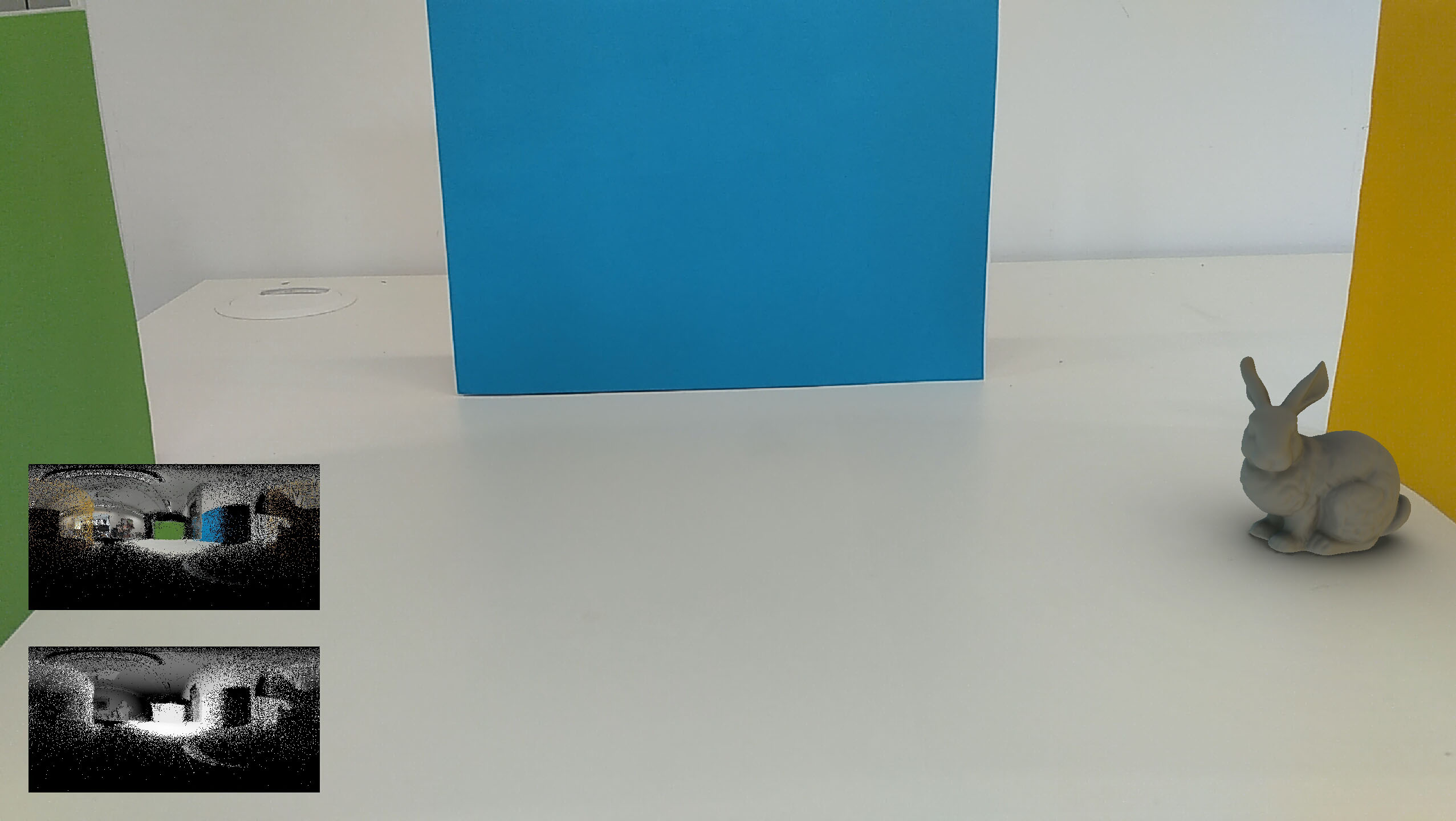}}
    \vspace{-5pt}
    \caption{Example of augmented scene when translating the EM. \textit{Left}: Original position. \textit{Right}: After translation. The corresponding EM (Colour + Depth Map) can be seen on the lower-left corners.}
    \label{fig:resMotion}
    \vspace{-15pt}
\end{figure}

\begin{figure*}
    \centering
    \subfigure{\includegraphics[width = 0.4\textwidth]{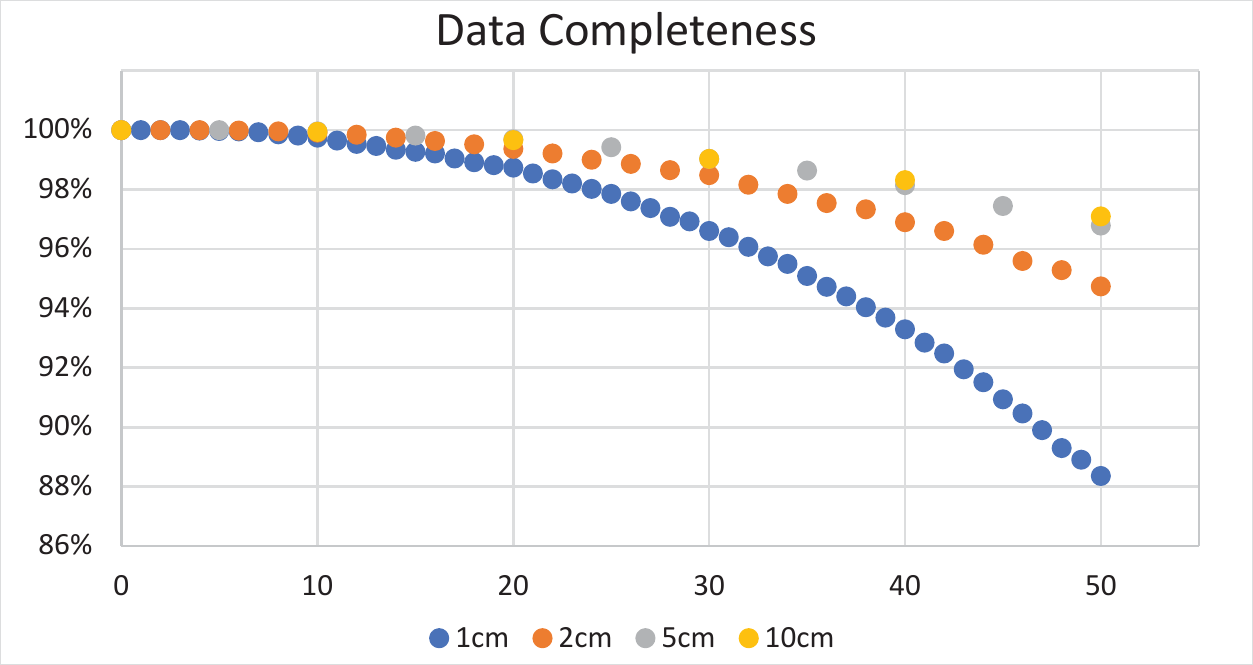}}
    \hspace{10pt}
    \subfigure{\includegraphics[width = 0.4\textwidth]{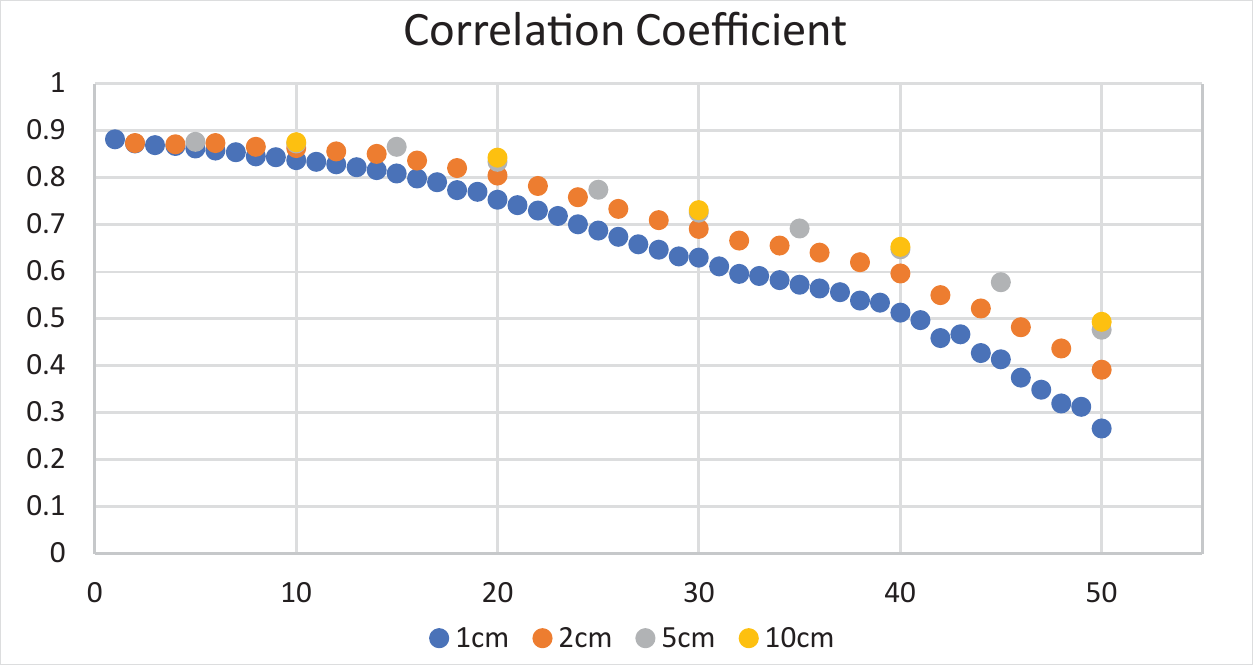}}
    \vspace{-5pt}
    \caption{\change{Data loss and decreasing correlation coefficient after continuous EM translations in steps of 1cm, 2cm, 5cm and 10cm.}}
    \label{fig:plots}
    \vspace{-15pt}
\end{figure*} 

\begin{figure}
    \centering
    \includegraphics[width=0.45\textwidth]{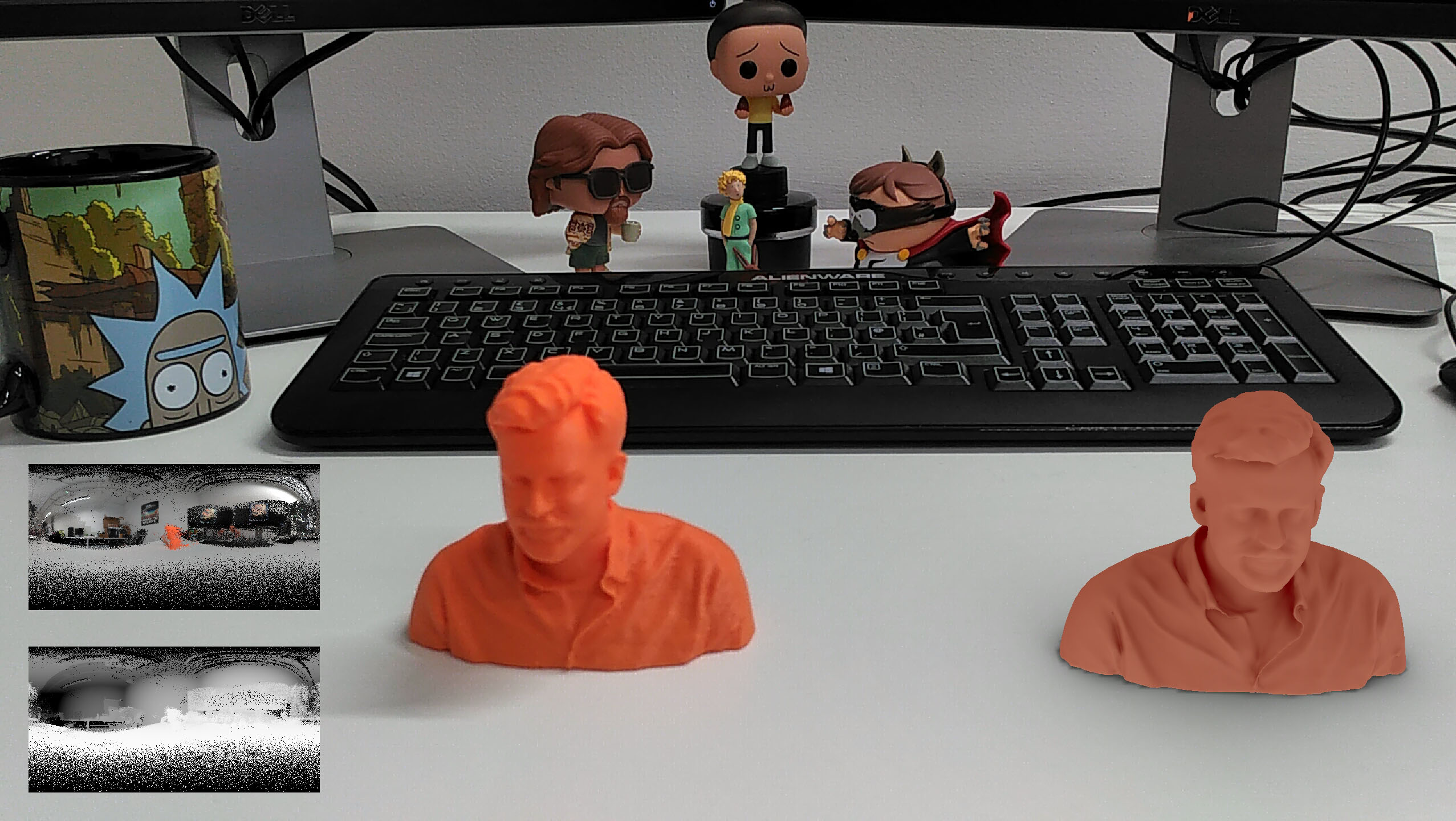}
    \caption{Comparison with a real object.}
    \label{fig:realObject}
    \vspace{-15pt}
\end{figure}

\change{\textbf{EM Translation} As described in Section~\ref{sec:EnvironmentCapture}, it is possible to translate the EM once it has been captured. We compared our method to the operator proposed by Gardner et al.~\cite{Gardner:2017:LPII}. A displacement of 10cm along the x-axis with both methods is shown in Fig.~\ref{fig:transComparison}. As can be seen, their operator overemphasises translations due to the assumption that all surfaces are far away. \\
Once the EM has been translated, the effects of the new position are visible on the virtual object.} Fig.~\ref{fig:resMotion} illustrates this situation. On the lower-left corners of each image, the corresponding EM is visualised: colour and depth data at the top and bottom respectively.\\
The scene consists of three coloured panels surrounding the area where the virtual object is located. As seen on the left side of Fig.~\ref{fig:resMotion}, once the EM has been acquired, the virtual object is rendered with its shadows and the effects due to the environment's radiance. When the virtual object is moved closer to one of the coloured panels, its effects on the object are visible: a yellow colour cast appears on its right side. As mentioned in Section~\ref{sec:transEM}, the EM can be relocated to the new position on a permanent basis, such that the device can restart the environment acquisition to fill the missing portions of the EM. The temporary relocation of the EM and its immediate effects are demonstrated in the supplementary video.\\
\change{If the EM is continuously translated without using temporary EMs, data will progressively be lost and its reliability compromised. Fig.~\ref{fig:plots} shows the effects on the available data after translating the EM 50cm in steps of 1cm, 2cm, 5cm and 10cm. The comparison is made with ground truth, defined as an EM that is fed the same RGB-D data but with a displaced origin matching that of the translated EM. Even if large displacements do not seem to lead to substantial losses of data, when analysed in terms of their correlation coefficient with respect to ground truth, it is clear that the effect is considerable. The cross correlation is calculated using the techniques proposed by Guti\'{e}rrez et al.~\cite{Gutierrez:2018:TDD}, designed for Omni-directional Images (ODIs), which use the same projection as our EMs. The average of the correlation coefficients per channel is here reported. As can be seen from both plots, the amount of information lost in a large displacement is smaller than the sum of the losses using smaller deltas to match the same translation. The EM translation is an expensive operation since it reprojects all pixels in the EM and takes on average 28.02ms. For visual comparisons of the translated EMs and their corresponding ground truth, please see the supplemental material.}\\ \\
We compared the final rendering with a 3D printed object. A sample case can be seen in Fig.~\ref{fig:realObject}. Even though hard shadows are not possible to generate with SH, the created soft shadow gives a plausible impression on the virtual object. 
\subsection{Limitations}
The results presented here demonstrate that our system is capable of interactively acquiring an environment radiance while rendering virtual objects at the same time. All within the current limitations of a mobile device. There are, however, some limitations due to the nature of the processes involved in the pipeline.\\
With regards to the environment capture, we are limited by the size of the EM and consequently very large rooms cannot be properly captured. This is due to the fact that surfaces far away from the EM's origin will be mapped to a small number of pixels and obtaining a proper colour correction matrix might not be feasible. \\
The colour correction algorithm assumes that all surfaces are perfectly Lambertian and without specularities. Acquiring an environment with shiny surfaces will face difficulties when estimating the correction matrix, leading to either incomplete EMs, or forcing the user to move the device around the surface until an appropriate correction matrix is obtained.\\
Even though our system is able to handle dynamic scenes, significant changes in the scene could interfere in the process of finding reliable paired samples to calculate a good correction matrix. One strategy to alleviate this, is to capture the area that has changed in increments, such that enough reliable samples are made available. The same strategy can also be performed in environments with light-emitting surfaces that trigger a strong white-balance adjustment, e.g., computer screens.\\
As shown in Fig.~\ref{fig:plots}, EM displacements cause a progressive loss of information on the updated EM. This restricts the magnitude of the motions that lead to usable SH coefficients allowing plausible environment interactions with the virtual object. \\
Since soft shadows are rendered using a static plane, interactions with nearby objects in the real-world are not possible. A solution for this restriction could be the one proposed by Sloan et al.~\cite{Sloan:2002:PRT}, where a series of SH coefficients are precomputed on a 3D grid around the virtual object.

\section{Conclusion and Future Work}
We have presented a novel system that allows for simultaneous environment acquisition and AR rendering which runs, on average, in real-time and well within the limitations of a mobile device. The proposed system takes as input RGB-D images and combines them into a 2D EM suitable to augment a real-world scene with coherently illuminated virtual objects. We demonstrated that our system achieves real-time augmentations on mobile devices.\\
The RGB-D EM, together with the set of defined criteria used to update it, permits a dynamic response when small changes occur in the scene. \change{Additionally, using the depth information in the EM, it is possible to simulate the effects of the environment when the virtual object is moved.} \\
On the Lenovo Phab 2 Pro, our system runs on average in real-time (31Hz) using 25 spherical harmonic coefficients, whereas recent state-of-the-art barely achieves interactive frame rates \cite{Rohmer:2017:NEI}. We demonstrated the quality of our renderings and the accuracy of the illumination estimation with qualitative results comparing a rendered model versus its real 3D printed version in real lighting conditions. \change{Specular effects are not considered at the moment, but techniques like that presented by K\'an et al.~\cite{Kan:2015:HQC} could be used as an extension to our system. Furthermore, with the advent of Deep Learning, we envision future research leading to systems capable of predicting plausible EMs from video streams or single images for all materials and situations. Meta et al.~\cite{Meka:2018:LIME} demonstrated this is possible for objects with simple materials.} \\
Even though the proposed pipeline is currently implemented using a Tango-enabled device, these techniques could be applied on any system with SLAM capabilities and able to provide RGB-D data.

\vspace{-5pt}

\acknowledgments{
This publication has emanated from research conducted with the financial support of Science Foundation Ireland (SFI) under the Grant Number 15/RP/2776.}

\bibliographystyle{eg-alpha-doi}

\vspace{-5pt}

\bibliography{references}

\end{document}